\documentclass{article}
\usepackage{spconf,amsmath,graphicx}
\usepackage{amsfonts}
\usepackage{multirow}
\usepackage{multicol}
\usepackage{booktabs}
\title{DPANet: Dual Pyramid Attention Network for Multivariate Time Series Forecasting}
\name{Qianyang Li$^{*1}$ \thanks{*Corresponding author: Qianyang Li and Xingjun Zhang. This work is partially supported by National Natural Science Foundation of China 62372366.}, Xingjun Zhang$^{*1}$,  Shaoxun Wang$^{1}$,Wei Jia$^{2}$ }

\address{
    $^{1}$School of Computer Science and Technology, Xi'an Jiaotong University, Xi'an, China\\
    $^{2}$Department of Computer Science and Technology, Tsinghua University, Beijing, China\\
    \{liqianyang,shaoxunwang\}@stu.xjtu.edu.cn, xjzhang@xjtu.edu.cn,  weijia4473@mail.tsinghua.edu.cn    
}
\begin{document}
%\ninept
%
\maketitle
\begin{abstract}
Long-term time series forecasting (LTSF) is hampered by the challenge of modeling complex dependencies that span multiple temporal scales and frequency resolutions. Existing methods, including Transformer and MLP-based models, often struggle to capture these intertwined characteristics in a unified and structured manner. We propose the Dual Pyramid Attention Network (DPANet), a novel architecture that explicitly decouples and concurrently models temporal multi-scale dynamics and spectral multi-resolution periodicities. DPANet constructs two parallel pyramids: a Temporal Pyramid built on progressive downsampling, and a Frequency Pyramid built on band-pass filtering. The core of our model is the Cross-Pyramid Fusion Block, which facilitates deep, interactive information exchange between corresponding pyramid levels via cross-attention. This fusion proceeds in a coarse-to-fine hierarchy, enabling global context to guide local representation learning. Extensive experiments on public benchmarks show that DPANet achieves state-of-the-art performance, significantly outperforming prior models. Code is available at https://github.com/hit636/DPANet.
\end{abstract}
\begin{keywords}
Time Series Forecasting, Attention Mechanism, Multi-Scale Learning, Pyramid Network.
\end{keywords}
\section{Introduction}
\label{sec:intro}

Time series forecasting is a critical task in numerous real-world applications, driven by its broad practical utility and complex theoretical questions \cite{kim2025comprehensive}. The field offers significant impacts across numerous critical sectors, including financial market analysis \cite{zhu2024lsr}, energy demand management \cite{deb2017review}, meteorological forecasting \cite{qiu2024tfb}, and transportation system optimization \cite{fang2025efficient}.While recent advances in deep learning, particularly with Transformer and MLP-based architectures \cite{zeng2023transformers}, have significantly improved performance, a fundamental challenge remains: effectively modeling the hierarchical structure inherent in time series data.

Real-world time series are compositions of multi-scale temporal patterns (e.g., daily vs. yearly trends) and multi-resolution frequency components (e.g., high-frequency fluctuations vs. low-frequency seasonalities). These properties are orthogonal yet deeply coupled; for instance, a long-term seasonal trend modulates high-frequency daily patterns. However, existing models like PatchTST\cite{nie2022time}, while excelling at long-range dependencies, lack an explicit structure for this intricate multi-resolution analysis.

This paper argues that addressing the challenges of Long-Term Time Series Forecasting (LTSF) requires a model architecture that mirrors the inherent multi-level structure of time series data. Drawing inspiration from pyramid networks in signal and image processing \cite{lin2017feature}, and building on recent advances in multi-scale fusion models \cite{wang2024timemixer}, we propose a fusion approach that decouples temporal and frequency hierarchies and integrates them through interactive fusion. We introduce the Dual Pyramid Attention Network (DPANet), which employs a dual-stream architecture with two pyramids: a Temporal Pyramid that captures multi-scale hierarchical features via downsampling, and a Frequency Pyramid that disentangles periodic patterns using band-pass filtering. The core innovation is the Cross-Pyramid Fusion Block, enabling deep information exchange between corresponding pyramid levels through cross-attention.

 Our contributions are threefold: (1) We propose a novel dual-pyramid architecture for parallel time and frequency domain analysis. (2) We design a cross-pyramid fusion mechanism for a deep hierarchical feature interaction. (3) We empirically establish the superiority of DPANet as a new strong baseline for LTSF.

\section{Related Work}
\label{sec:relate}

\subsection{Transformer-based Time Series Forecasting}
Transformer-based models have set the state of the art by leveraging attention to capture long-range dependencies. Innovations have focused on improving efficiency and adapting the mechanism for temporal data. Notable examples include Informer \cite{zhou2021informer} with its ProbSparse attention, Autoformer \cite{wu2021autoformer} with its auto-correlation mechanism, and FEDformer \cite{zhou2022fedformer} which operates in the frequency domain. More recently, PatchTST \cite{nie2022time}, inspired by Vision Transformers (ViT), has shown great success by processing time series in patches. While powerful, these methods typically operate within a single domain (either time or frequency). DPANet distinguishes itself by explicitly modeling both domains in parallel and facilitating deep, hierarchical interaction between them.
\subsection{MLP-based Time Series Forecasting}
Recent works have demonstrated that lightweight MLP-based models can be surprisingly effective. N-BEATS \cite{oreshkin2019n} used a deep stack of MLP blocks with a basis expansion principle. DLinear \cite{zeng2023transformers} further simplified this, showing that a single linear layer applied to decomposed trend and seasonality can outperform complex Transformers. Following this trend, models like TSMixer \cite{ekambaram2023tsmixer} have successfully adapted the MLP-Mixer architecture for efficient temporal and channel-wise feature extraction. DPANet learns from this paradigm by using efficient computational blocks but embeds them within a structured, hierarchical framework to explicitly capture the multi-scale and multi-resolution properties that simple MLPs overlook.
\subsection{Multi-Scale and Multi-Resolution Methods}
Multi-scale and multi-resolution analysis is a classic approach to time series that has been integrated into various deep learning models. Pyraformer \cite{liu2022pyraformer} used a pyramidal attention structure to reduce complexity, while TimesNet \cite{DBLP:conf/iclr/TimesNet} reshapes 1D series into 2D tensors based on periodicities discovered via FFT. Recent mixing models like TimeMixer \cite{wang2024timexer} leverage multi-scale decomposition. DPANet advances this line of research by constructing two distinct, parallel pyramids for the temporal and spectral domains and introduces a novel mechanism for their interactive fusion, enabling a more comprehensive and principled hierarchical analysis than prior works.

\section{Method}
\label{sec:method}

\subsection{DPANet Architecture}
Given a historical multivariate time series $\mathbf{X} \in \mathbb{R}^{L_{\text{in}} \times C}$, where $L_{\text{in}}$ is the look-back window length and $C$ is the number of variates, our objective is to predict the future sequence $\mathbf{Y} \in \mathbb{R}^{L_{\text{pred}} \times C}$ of length $L_{\text{pred}}$.

\begin{figure*}[htbp]
    % Replace 'model_architecture.pdf' with the actual path to your figure file
    \includegraphics[width=\textwidth]{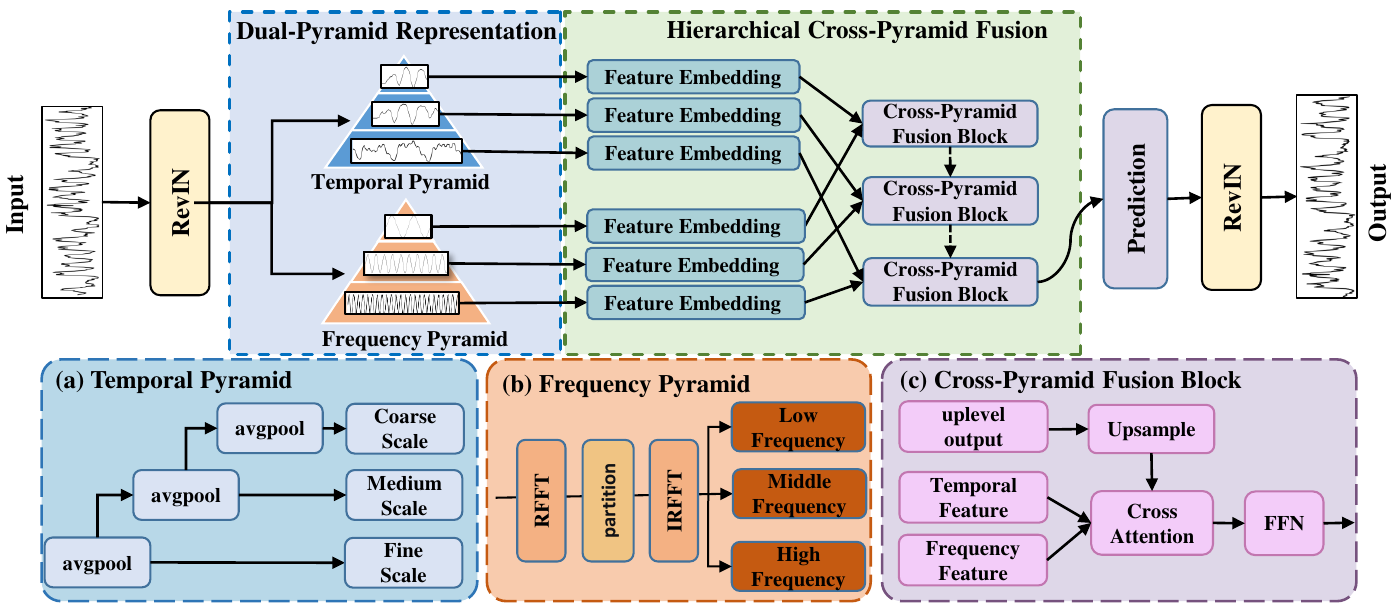} 
    \caption{The overall architecture of DPANet. The model decouples a time series into parallel (a) Temporal and (b) Frequency Pyramids. These representations are then fused through a stack of (c) Cross-Pyramid Fusion Blocks in a coarse-to-fine hierarchy.}
    \label{fig:model_arch}
\end{figure*}

As illustrated in Figure~\ref{fig:model_arch}, the DPANet architecture systematically decouples and fuses multi-scale temporal and multi-resolution information. The input is first normalized via a RevIN layer \cite{kim2021reversible}. The model's core is a stack of novel \textbf{Cross-Pyramid Fusion Blocks} operating on parallel temporal and frequency pyramids. The fusion adheres to a coarse-to-fine hierarchical principle, where high-level semantics guide low-level feature learning. A final prediction head maps the fused representation to the forecast, which is then de-normalized by the inverse RevIN transformation.

\subsection{Dual-Pyramid Representation}

To simultaneously capture dynamics across different time granularities and frequency bands, we construct two parallel feature pyramids from the normalized input $\mathbf{X}'$.

\textbf{Temporal Pyramid:}
The temporal pyramid, $\mathcal{P}_t = \{\mathbf{T}^{(0)}, \dots, \mathbf{T}^{(S-1)}\}$, is created by progressively downsampling the input series $\mathbf{X}'$ in the time domain. Let $\mathbf{T}^{(0)} = \mathbf{X}'$. Each subsequent level is obtained by:
\begin{equation}
\mathbf{T}^{(s)} = \text{AvgPool1d}(\mathbf{T}^{(s-1)}) \in \mathbb{R}^{(L_{\text{in}}/2^s) \times C}
\end{equation}
where $s \in \{1, \dots, S-1\}$ is the scale level. This pyramid effectively captures the series' structure from fine-grained details to coarse-grained trends.

\textbf{Frequency Pyramid:} The frequency pyramid $P_f = \{F^{(0)}, \dots, F^{(S-1)}\}$ is created by decomposing the input series into different frequency resolutions. 
We first compute the Real Fourier Transform of the input $\mathcal{F} = \text{RFFT}(X')$. 
The spectrum is then partitioned into $S$ disjoint, logarithmically-spaced bands. 
This logarithmic partitioning is crucial: it allocates finer resolution to the low-frequency components, which often correspond to the dominant, long-term seasonalities, while grouping high-frequency noise more coarsely.

For each band $s$, we create a binary mask $M^{(s)}$ to isolate its range of frequencies. The corresponding pyramid level $F^{(s)}$ is reconstructed by applying this mask and transforming the result back into the time domain via the Inverse RFFT:
\begin{equation}
    F^{(s)} = \text{IRFFT}(\mathcal{F} \odot M^{(s)}) \in \mathbb{R}^{L_{in} \times C}
    \label{eq:freq_pyramid}
\end{equation}
where $\odot$ denotes element-wise multiplication. Each $F^{(s)}$  represents a specific periodic component of the original series, effectively isolating different seasonal patterns for the fusion stage.

\subsection{Hierarchical Cross-Pyramid Fusion}

The core of DPANet is the hierarchical fusion process, which operates from the coarsest scale to the finest. At each scale $s$, a Cross-Pyramid Fusion Block is employed to facilitate deep interaction between the temporal and frequency representations.

\textbf{Feature Embedding:} Before fusion, the representations from each pyramid, $\mathbf{T}^{(s)}$ and $\mathbf{F}^{(s)}$, are first reshaped to treat each channel as an independent instance, and then embedded into a $d_\text{model}$-dimensional feature space via scale-specific linear layers, yielding initial hidden states $\mathbf{h}_t^{(s)}$ and $\mathbf{h}_f^{(s)}$.This reshaping treats each channel's time series as an independent instance for the initial embedding, while their relationships are later captured by the cross-attention mechanism.

\textbf{Cross-Pyramid Fusion Block:} This block enables bidirectional information flow between the two domains. Let $\mathbf{h}_t$ and $\mathbf{h}_f$ be the input representations  $\mathbf{h}_t^{(s)}$ and $\mathbf{h}_f^{(s)}$ for the current scale. First, two parallel cross-attention layers enrich the representations:
\begin{align}
    \mathbf{h}_t' &= \text{LayerNorm}(\mathbf{h}_t + \text{CrossAttn}(\mathbf{h}_t, \mathbf{h}_f, \mathbf{h}_f)) \\
    \mathbf{h}_f' &= \text{LayerNorm}(\mathbf{h}_f + \text{CrossAttn}(\mathbf{h}_f, \mathbf{h}_t, \mathbf{h}_t))
\end{align}
where $\text{CrossAttn}(Q, K, V)$ denotes a standard multi-head attention module. The two enriched representations are then concatenated and processed by a Feed-Forward Network (FFN) for deep fusion. The output is finally split back into the updated temporal and frequency representations, $\tilde{\mathbf{h}}_t^{(s)}$ and $\tilde{\mathbf{h}}_f^{(s)}$.

The fusion process starts at the coarsest scale ($s=S-1$) and proceeds to the finest ($s=0$). For each finer scale $s$, the fused outputs from the previous coarser scale, $\bar{h}_t^{(s+1)}$ and $\bar{h}_f^{(s+1)}$, are upsampled via \textbf{linear interpolation} to match the current sequence length. They are then added as a residual connection to the initial embeddings, $h_t^{(s)}$ and $h_f^{(s)}$, creating the inputs for the current fusion block:
\begin{align}
h_t^{(s)'} &= h_t^{(s)} + \text{UpSample}(\bar{h}_t^{(s+1)}) \label{eq:res_t} \\
h_f^{(s)'} &= h_f^{(s)} + \text{UpSample}(\bar{h}_f^{(s+1)}) \label{eq:res_f}
\end{align}
where $\text{UpSample}(\cdot)$ doubles the sequence length. This coarse-to-fine mechanism allows abstract, global information to guide representation learning at finer levels.
\subsection{Prediction} 

The final forecast is generated from the temporal representation $\hat{h}^{(0)}$ at the finest scale ($s=0$), which contains the fully integrated multi-scale and multi-resolution information. A linear prediction head first projects a sequence-pooled representation of $\hat{h}^{(0)}$ to the desired prediction length, $L_{pred}$, yielding a raw forecast in the normalized space. Finally, this raw forecast is de-normalized into the final output by applying the inverse RevIN transformation with the stored input statistics.

\section{Experiments}
\label{sec:experiment}

\subsection{Datasets and Baselines}
\label{ssec:dataset}

We conducted experiments on 8 widely-used LTSF benchmarks: ETT (ETTH1,ETTh2, ETTm1, ETTm2), Weather,Electricity, and Traffic. We compared DPANet with a comprehensive set of SOTA baselines, including Transformer-based models (PatchTST,TimeXer), MLP-based models (DLinear, TSMixer), Multiscale models (TimeMixer,MSD-Mixer,TimesNet), and other pyramid-based models (Pyraformer).

\subsection{Experimental Setup}
\label{sssec:subsubhead}

We followed the standard evaluation protocol,  all models were evaluated using a fixed input sequence length of $L=96$ across four prediction horizons ${T \in \{96, 192, 336, 720\}}$. All models were trained under identical conditions with fair hyperparameter tuning. Mean Squared Error (MSE) and Mean Absolute Error (MAE) were used as evaluation metrics.
\begin{table*}[t]
\caption{Comprehensive results for the long-term forecasting task.The best results are highlighted in \textbf{bold}, while the second-best results are \underline{underlined}.}
\centering
\resizebox{\textwidth}{!}{%
\begin{tabular}{ll  rr rr rr rr rr rr rr rr rr  }
\toprule
\multicolumn{2}{l}{\textbf{Models}} & \multicolumn{2}{c}{\textbf{DPANet}} & \multicolumn{2}{c}{\textbf{TimeMixer}}  &\multicolumn{2}{c} {\textbf{MSD-Mixer}} & \multicolumn{2}{c}{\textbf{TimeXer}} & \multicolumn{2}{c}{\textbf{PatchTST}}  & \multicolumn{2}{c}{\textbf{TimesNet}} & \multicolumn{2}{c}{\textbf{DLinear}} & \multicolumn{2}{c}{\textbf{FEDformer}} & \multicolumn{2}{c}{\textbf{Pyraformer}} \\
% \multicolumn{2}{l}{} & \multicolumn{2}{c}{\textbf{(Ours)}} & \multicolumn{2}{c}{\textbf{2024b}} & \multicolumn{2}{c}{\textbf{2024}}  & \multicolumn{2}{c}{\textbf{2024}} & \multicolumn{2}{c}{\textbf{2024}}& \multicolumn{2}{c}{\textbf{2023}}  & \multicolumn{2}{c}{\textbf{2023}} & \multicolumn{2}{c}{\textbf{2023}} & \multicolumn{2}{c}{\textbf{2022}} & \multicolumn{2}{c}{\textbf{2021}} \\
\cmidrule(lr){3-4} \cmidrule(lr){5-6} \cmidrule(lr){7-8} \cmidrule(lr){9-10} \cmidrule(lr){11-12} \cmidrule(lr){13-14} \cmidrule(lr){15-16} \cmidrule(lr){17-18} \cmidrule(lr){19-20} 
\textbf{Metric} & & \textbf{MSE} & \textbf{MAE} & \textbf{MSE} & \textbf{MAE} & \textbf{MSE} & \textbf{MAE} & \textbf{MSE} & \textbf{MAE} & \textbf{MSE} & \textbf{MAE} & \textbf{MSE} & \textbf{MAE} & \textbf{MSE} & \textbf{MAE} & \textbf{MSE} & \textbf{MAE} & \textbf{MSE} & \textbf{MAE}   \\

\midrule
\multirow{5}{*}{\rotatebox{90}{ETTh1}} & 96 & \textbf{0.371} & \underline{0.398} & \underline{0.375} & {0.400} & 0.377 & \textbf{0.391} & 0.382 & 0.403 & 0.460 & 0.447  & 0.384 & 0.402 & 0.407 & 0.412 & 0.395 & 0.424 & 0.664 & 0.612  \\
& 192 & \textbf{0.420} & \textbf{0.419} & {0.429} & \underline{0.421} & \underline{0.427} & 0.422  & 0.429 & 0.453 & 0.477 & 0.429  & 0.436 & 0.429 & 0.446 & 0.441 & 0.469 & 0.470 & 0.790 & 0.681  \\
& 336 & {\textbf{0.441}} & \textbf{0.441} & \underline{0.458} & {0.448} & 0.469 & \underline{0.443} &  0.468 & 0.448 & 0.546 & 0.496  & 0.491 & 0.469 & 0.489 & 0.467 & 0.547 & 0.495 & 0.891 & 0.738  \\
& 720 & \textbf{0.461} & \textbf{0.458} & 0.498 & 0.482 & 0.485 & 0.475 & \underline{0.469} & \underline{0.461} & 0.544 & 0.517  & 0.521 & 0.500 & 0.513 & 0.510 & 0.598 & 0.544 & 0.963 & 0.782   \\

\midrule
\multirow{5}{*}{\rotatebox{90}{ETTh2}} & 96 & {0.292}  & \textbf{0.340} & \underline{0.289} & \underline{0.342} & \textbf{0.284} & 0.345  & 0.308 & 0.355 & 0.745 & 0.584  & 0.340 & 0.374 & 0.340 & 0.394 & 0.358 & 0.397 & 0.645 & 0.597  \\
& 192 & {0.374} & \textbf{0.391} & {0.378} & {0.397} & \textbf{0.362} &\underline{0.392}  &\underline{0.363} & 0.389 & 0.393 & 0.405  & 0.402 & 0.414 & 0.482 & 0.479 & 0.429 & 0.439  & 0.788 & 0.683  \\
& 336 & \textbf{0.383} & \textbf{0.411} & \underline{0.386} & \underline{0.414}& 0.399 & 0.428  & 0.414 & 0.423 & 0.427 & 0.436  & 0.452 & 0.452 & 0.591 & 0.541 & 0.496 & 0.487 & 0.907 & 0.747   \\
& 720 & \textbf{0.411} & \textbf{0.430} & \underline{0.412} & {0.434} & 0.426 & 0.457  & {0.414} & \underline{0.432} & 0.436 & 0.450  & 0.462 & 0.468 & 0.839 & 0.661 & 0.463 & 0.474 & 0.963 & 0.783  \\

\midrule
\multirow{5}{*}{\rotatebox{90}{ETTm1}} & 96 & \underline{0.319} & \textbf{0.349} & 0.320 & 0.357 & \textbf{0.304} & \underline{0.351} & {0.318} & {0.356} & 0.352 & 0.374  & 0.338 & 0.375 & 0.346 & 0.374 & 0.379 & 0.419 & 0.543 & 0.510  \\
& 192 & \textbf{0.343} & \underline{0.378} & {0.361} & {0.381} & \underline{0.344}  & \textbf{0.375}  & 0.362 & 0.383 & 0.374 & 0.387  & 0.374 & 0.387 & 0.382 & 0.391 & 0.389 & 0.387 & 0.557 & 0.537  \\
& 336 & \textbf{0.379} & \textbf{0.393} & {0.390} & {0.404} & \underline{0.380} & \underline{0.395}  & 0.395 & 0.407 & 0.421 & 0.414  & 0.410 & 0.411 & 0.415 & 0.415 & 0.445 & 0.459 & 0.754 & 0.655   \\
& 720 & \underline{0.431} & \textbf{0.422} & 0.454 & 0.441 & \textbf{0.427} & \underline{0.428}  & {0.452} & {0.441} & 0.462 & 0.449  & 0.478 & 0.450 & 0.473 & 0.451 & 0.543 & 0.490 & 0.908 & 0.724  \\

\midrule
\multirow{5}{*}{\rotatebox{90}{ETTm2}} & 96 & \underline{0.173} & \textbf{0.255} & 0.175 & 0.258 & \textbf{0.169} & 0.259 & {0.171} & \underline{0.256} & 0.183 & 0.270  & 0.187 & 0.267 & 0.193 & 0.293 &  0.203 & 0.287  & 0.435 & 0.507  \\
& 192 & \textbf{0.233} & \textbf{0.297} & {0.237} & \underline{0.299} & \underline{0.232} & 0.300  & 0.237 & 0.299 & 0.255 & 0.314  & 0.249 & 0.309 & 0.284 & 0.361 & 0.269 & 0.328 & 0.730 & 0.673  \\
& 336 & \textbf{0.291} & \textbf{0.335} & 0.298 & 0.340  & \underline{0.292} & \underline{0.337} & {0.296} & {0.338} & 0.309 & 0.347 & 0.321 & 0.351 & 0.382 & 0.429 & 0.325 & 0.366 & 1.201 & 0.845  \\
& 720 & \textbf{0.390} & \underline{0.395} & \underline{0.391} & {0.396} & 0.392 & 0.398 & 0.392 & \textbf{0.394} & 0.412 & 0.404  & 0.408 & 0.403 & 0.558 & 0.525 & 0.421 & 0.415  & 3.345 & 1.451   \\

\midrule
\multirow{5}{*}{\rotatebox{90}{Weather}} & 96 &  {0.168} &  \underline{0.211}& 0.202 & 0.261  & \textbf{0.148} & 0.212& \underline{0.157} & \textbf{0.205} & 0.186 & 0.227 & {0.172} & {0.220} & 0.195 & 0.252 & 0.217 & 0.296  & 0.622 & 0.566   \\
& 192 & \underline{0.207} & \textbf{0.246} & 0.208 & \underline{0.250} & \textbf{0.200} & 0.262  & {0.204} & \underline{0.247} & 0.234 & 0.265  & 0.219 & 0.261 & 0.237 & 0.295 & 0.276 & 0.336 & 0.739 & 0.624   \\
& 336 & \textbf{0.250} & \underline{0.285} & \underline{0.251} & \textbf{0.278} & 0.256 & 0.310  & 0.261 & 0.290 & 0.284 & 0.301  & 0.280 & 0.306 & 0.282 & 0.331 & 0.339 & 0.380 & 1.004 & 0.753   \\
& 720 & \textbf{0.338} & \textbf{0.339} & \underline{0.339} &\underline{0.341} & 0.327 & 0.362  & {0.340} & 0.341 & 0.356 & 0.349  & 0.365 & 0.359 & 0.359 & {0.345} & 0.403 & 0.428 & 1.420 & 0.934  \\

\midrule
\multirow{5}{*}{\rotatebox{90}{Electricity}} & 96 & \textbf{0.152} & \textbf{0.245} &  \underline{0.153} &  \underline{0.247}  & 0.161 & 0.253  & 0.182 & 0.278 & 0.190 & 0.296 & 0.168 & 0.272 & 0.210 & 0.305 & 0.169 & 0.273 & 0.386 & 0.449   \\
& 192 & \textbf{0.161} & \underline{0.257} & \underline{0.166} & \textbf{0.256} & 0.169 & 0.269  & 0.199 & 0.263 & 0.196 & 0.304   & 0.184 & 0.322 & 0.210 & 0.305 & 0.201 & 0.315 & 0.378 & 0.443  \\
& 336 & \textbf{0.183} & \textbf{0.273} &\underline{0.185} & \underline{0.277} & 0.188 & 0.286  & 0.193 & 0.312 & 0.217 & 0.319  & 0.198 & 0.300 & 0.223 & 0.319 & 0.200 & 0.304 & 0.376 & 0.443   \\
& 720 & \underline{0.227} & \textbf{0.307} & \textbf{0.225} & 0.317 & {0.229}  & \underline{0.311} & 0.233 & 0.312 & 0.258 & 0.352 & 0.220 & 0.320 & 0.258 & 0.350 & 0.246 & 0.355 & 0.376 & 0.445 \\

\midrule
\multirow{5}{*}{\rotatebox{90}{Traffic}} & 96 & \underline{0.459} & \underline{0.274} & 0.462 & 0.285 & 0.500 & 0.324  & \textbf{0.428} & \textbf{0.271} & 0.526 & 0.347 & 0.593 & 0.321 & 0.650 & 0.396 & 0.587 & 0.366  & 0.867 & 0.468  \\
& 192 &  \underline{0.471} & \underline{0.291} & {0.473} & {0.296} & 0.500 & 0.324  & \textbf{0.448} & \textbf{0.282} & 0.522 & 0.332  & 0.617 & 0.336 & 0.598 & 0.370 & 0.604 & 0.373 & 0.869 & 0.467  \\
& 336 & \underline{0.492} &  \underline{0.295} & 0.498 & {0.296} & 0.528 & 0.341 & \textbf{0.473} & \textbf{0.289} & 0.517 & 0.334 & 0.629 & 0.336 & 0.605 & 0.373 & 0.621 & 0.383  & 0.881 & 0.469  \\
& 720 & \textbf{0.505} & \textbf{0.306} & \underline{0.506} & 0.313 & 0.561 & 0.369& 0.516 & \underline{ 0.307} & 0.552 & 0.352   & 0.640 & 0.350 & 0.645 & 0.394 & 0.626 & 0.382  & 0.896 & 0.473   \\

\midrule
\multicolumn{2}{l}{1st Count} & \textbf{17} & \textbf{19} & 1 & 2 & 7 & 2  & 3 & 5 & 0 & 0 & 0 & 0 & 0 & 0 & 0 & 0 & 0 & 0  \\
\bottomrule
\end{tabular}
}

\label{tab:full_long_term_forecasting}
\end{table*}

\subsection{Main Results}
\label{sec:result}

As shown in Table \ref{tab:full_long_term_forecasting}, our model consistently demonstrates superior performance across seven benchmarks. DPANet achieves the best results in 17 MSE and 19 MAE metrics out of 28 total settings. Its advantage is particularly pronounced on datasets with complex patterns like Electricity and ETTh1, where DPANet secures the top rank across all prediction horizons for both MSE and MAE. Furthermore, on the highly volatile Traffic dataset, DPANet proves its robustness by outperforming strong competitors in the most challenging long-range forecast (horizon 720). These results validate the efficacy of our dual-pyramid architecture.

\subsection{Ablation Study}
\label{sec:ablation}
We conducted rigorous ablation studies to validate DPANet's key components (Table \ref{tab:ablation-study}). The full model consistently outperforms all variants. To test our dual-domain hypothesis, we designed two specialized versions: a Temporal-Only model (fusing two identical temporal pyramids) and a Frequency-Only model (fusing two spectral pyramids). Both variants underperformed significantly, confirming that the fusion of heterogeneous temporal and frequency information is critical. Furthermore, replacing the cross-attention mechanism with a simpler method (w/o Cross-Fusion) caused the most severe performance degradation. This result underscores that our interactive fusion block is the most essential component.
\begin{table}[t]
\caption{Ablation study of DPANet components on the ETTm2 and Weather datasets.}
\centering
\resizebox{1.1\linewidth}{!}{% To make the table fit into the page width
\setlength{\tabcolsep}{4pt}
\begin{tabular}{llcccccccc}
\toprule
\multicolumn{2}{c}{\textbf{Models}} & \multicolumn{2}{c}{\textbf{DPANet}} & \multicolumn{2}{c}{\textbf{Temporal-Only}} & \multicolumn{2}{c}{\textbf{Frequency-Only}} & \multicolumn{2}{c}{\textbf{w/o Cross-Fusion}}  \\
\cmidrule(lr){3-4} \cmidrule(lr){5-6} \cmidrule(lr){7-8} \cmidrule(lr){9-10}  
\textbf{Metric} & & MSE & MAE & MSE & MAE & MSE & MAE & MSE & MAE \\
\midrule
\multirow{4}{*}{Ettm2} 
& 96  & \textbf{0.173} & \textbf{0.255} & 0.182 & 0.276 & 0.186 & 0.259 & 0.215 & 0.295   \\
& 192 & \textbf{0.233} & \textbf{0.297} & 0.242 & 0.325 & 0.240 & 0.319 & 0.268 & 0.412   \\
& 336 & \textbf{0.291} & \textbf{0.335} & 0.316 & 0.362 & 0.309 & 0.355 & 0.342 & 0.468   \\
& 720 & \textbf{0.390} & \textbf{0.395} & 0.412 & 0.417 & 0.426 & 0.409 & 0.461 & 0.472   \\

\midrule
\multirow{4}{*}{Weather} 
& 96  & \textbf{0.168} & \textbf{0.211} & 0.179 & 0.223 & 0.176 & 0.218 & 0.192 & 0.253\\
& 192 & \textbf{0.207} & \textbf{0.246} & 0.220 & 0.257 & 0.223 & 0.242 & 0.243 & 0.293  \\
& 336 & \textbf{0.250} & \textbf{0.285} & 0.268 & 0.312 & 0.261 & 0.304 & 0.268 & 0.332  \\
& 720 & \textbf{0.338} & \textbf{0.339} & 0.351 & 0.352 & 0.349 & 0.346 & 0.373 & 0.412  \\

\bottomrule
\end{tabular}
}

\label{tab:ablation-study}
\end{table}

\section{Conclusion}
\label{sec:page}

In this paper, we introduced the Dual Pyramid Attention Network (DPANet) to address the challenge of modeling intertwined multi-scale and multi-resolution patterns in LTSF. By constructing parallel temporal and frequency pyramids and fusing them via a principled, coarse-to-fine cross-attention mechanism, DPANet achieves a more comprehensive and structured representation of complex time series dynamics. Extensive experimental results demonstrate that DPANet achieves superior performance across multiple benchmarks.

\vfill\pagebreak


\bibliographystyle{IEEEbib}
\bibliography{Template}

\end{document}